\documentclass{article}

\usepackage{subfiles}

\usepackage[preprint]{neurips_2022}

\usepackage[utf8]{inputenc} 
\usepackage[T1]{fontenc}    
\usepackage{hyperref}       
\usepackage{url}            
\usepackage{booktabs}       
\usepackage{amsfonts}       
\usepackage{nicefrac}       
\usepackage{microtype}      
\usepackage{xcolor}         
\usepackage{amsthm}
\usepackage{amsmath,amssymb}

\usepackage{graphicx}
\usepackage{subfigure}
\usepackage{parskip}
\usepackage{tikz}
\usepackage{lipsum}
\usepackage{bbm}
\usepackage{algorithm}
\usepackage{algorithmic}
\usepackage{amsmath}
\usepackage{multirow}
\usepackage{hhline}
\usepackage{hyperref}
\usepackage{mathtools}
\usepackage{csquotes}
\newtheorem{theorem}{Theorem}[section]

\newtheorem{lemma}[theorem]{Lemma}

\theoremstyle{definition}

\newtheorem{assumption}[theorem]{Assumption}

\newcommand{\E}{\mathbb{E}}
\newcommand{\X}{\textbf{X}}
\newcommand{\Xx}{\textbf{$\mathcal{X}$}}
\newcommand{\loss}{\mathcal{L}}
\newcommand{\W}{\mathcal{W}}
\newcommand{\F}{\mathcal{F}}
\newcommand{\D}{\textbf{\textit{D}}}

\newcommand{\bl}{\bigg(}
\newcommand{\br}{\bigg)}

\newcommand{\cka}{\text{Linear-CKA}}

\title{Federated Self-supervised Learning for Heterogeneous Clients}

\author{
  Disha Makhija \\
  University of Texas at Austin, \\
  \texttt{disham@utexas.edu}
   \And
   Nhat Ho \\
   University of Texas at Austin, \\
   \texttt{minhnhat@utexas.edu}
   \And 
   Joydeep Ghosh \\
   University of Texas at Austin, \\
   \texttt{jghosh@utexas.edu}
}

\begin{document}

\maketitle

\begin{abstract}
Federated Learning has become an important learning paradigm due to its privacy and computational benefits. As the field advances, two key challenges that still remain to be addressed are: (1) system heterogeneity - variability in the compute and/or data resources present on each client, and (2) lack of labeled data in certain federated settings. Several recent developments have tried to overcome these challenges independently. In this work, we propose a unified and systematic framework, \emph{Heterogeneous Self-supervised Federated Learning} (Hetero-SSFL) for enabling self-supervised learning with federation on heterogeneous clients. The proposed framework allows collaborative representation learning across all the clients without imposing architectural constraints or requiring presence of labeled data. The key idea in Hetero-SSFL is to let each client train its unique self-supervised model and enable the joint learning across clients by aligning the lower dimensional representations on a common dataset. The entire training procedure could be viewed as self and peer-supervised as both the local training and the alignment procedures do not require presence of any labeled data. As in conventional self-supervised learning, the obtained client models are task independent and can be used for varied end-tasks.
We provide a convergence guarantee of the proposed framework for non-convex objectives in heterogeneous settings and also empirically demonstrate that our proposed approach outperforms the state of the art methods by a significant margin. 


\end{abstract}
\let\thefootnote\relax\footnotetext{$\star$ Nhat Ho and Joydeep Ghosh contributed equally to this work. }

\section{Introduction}
Federated learning has become an important learning paradigm for training algorithms in a privacy preserving way and has gained a lot of interest in the recent past. While traditional federated learning is capable of learning high performing models~\citep{ditto}, two practical challenges that remain under studied are: system heterogeneity, and lack of labeled data. In several real world scenarios, the clients involved in the training process are highly heterogeneous in terms of their data and compute resources.  Requiring each client to train identical models like in traditional FL may thus be severely limiting. Similarly, assuming each client's local data resources to be fully labeled may not be pragmatic as annotating data is time consuming and may require expertise.  

Our focus in this work is to jointly address these two challenges in a systemic way so as to make FL more practical. Some prior works have studied these two issues separately. For instance, to help with scarcity of labeled data on local clients, use of semi-supervised learning has been recently  studied~\citep{semi-SSFL, semi-federated, semi-federated2};  other very recently proposed approaches suggest the use of locally self-supervised learning~\citep{FedU, divergenceaware, ssfl} assuming identical model architectures on each client. System heterogeneity on the other hand is commonly addressed by building heterogeneous personalised models on clients as suggested in~\citep{fedhenn, fedproto, jiang2020improving, perFL} but all of these approaches assume the presence of fully labeled data on all of the clients. 

Though the above mentioned approaches have shown promising results in dealing with both the problems in isolation,  in certain real-world applications these two challenges need to be jointly addressed in a holistic way. For example, some of the key applications of FL in cross-silo settings are in healthcare systems where different hospitals may possess varying amounts of private medical images or in cross-device settings where end-devices such as sensors acquire loads of unlabeled data. In both these cases, the data can neither be centralised nor can undergo extensive annotations, moreover to expect each client (a hospital or a sensor in the above examples) to train local models of identical capacities will be highly inhibiting. 

In this work, we circumvent these problems by proposing a new framework that allows clients to train  self-supervised models with unique (locally tuned) structures while still using the learnings from other clients in the network, in an architecture agnostic way. To achieve this, we add a proximal term to each client's loss function that helps in aligning the learnt lower dimensional representations (aka embeddings) across clients. This way of transferring and acquiring global knowledge allows variability in client model architectures while keeping the entire learning process independent of labeled training data. Furthermore, we use a kernel based distance metric for proximity calculation which provides much more flexibility to the clients in defining their own lower dimensional space without any constraints. To the best of our knowledge, this is the first work that studies collaborative unsupervised learning in a distributed heterogeneous client setting.

\textbf{Our Contributions} are summarized as follows :
\begin{enumerate}
    \item Our main contribution is the new framework, \emph{Hetero-SSFL}, for training heterogeneous models in a federated setting in an unsupervised way. Hetero-SSFL allows each client to train its own customized model architecture, using local data and computing resources while also utilising unlabeled supervision from peers.
    \item We perform extensive experimentation to provide a thorough evaluation of the approach in various kinds of practical settings. We observe that the proposed flexible approach substantially improves the accuracy of unsupervised federated learning in heterogeneous settings.
    \item We also provide theoretical analysis and convergence guarantee of the algorithm for non-convex loss functions.
\end{enumerate}

\textbf{Organization.} The rest of the paper is organised as follows. Section~\ref{sec:lit_review} provides a brief background on Federated Learning, Self-supervised Learning and related developments. In Section~\ref{sec:methodology}, we go over the preliminaries and then propose our framework. We study the convergence guarantee of the algorithm in Section~\ref{sec:analysis} and include the related proofs in the Appendix. A thorough experimental evaluation of our method on different types of datasets is presented in Section~\ref{sec:experiments} and we conclude the paper in Section~\ref{sec:discussion}.

\section{Related Work} \label{sec:lit_review}
This section provides an overview of the most relevant prior work in the fields of federated learning, self-supervised learning and federated self-supervised learning.

\subsection{Federated Learning(FL)}
The problem of training machine learning models in distributed environments with restrictions on data/model sharing was studied by several researchers in the data mining community in the early 2000s under titles such as (privacy preserving) distributed data mining~\citep{kapa00,gali17,aggarwal2008general}. Much of this work was for specific procedures such as distributed clustering~\citep{megh03} including in heterogenous settings with different feature spaces~\citep{megh05}, and  distributed PCA~\citep{kahu01}, or for specific models such as SVMs~\citep{yuji06}. Subsequently, with the re-surfacing in popularity of neural networks and proliferation of powerful deep learning approaches, the term \enquote{Federated Learning} got coined and popularized largely by an influential paper~\citep{pmlr-v54-mcmahan17a} which introduced FedAvg. Indeed FedAvg is now considered the standard distributed training method for Federated Learning, which has become a very active area of research since then. One of the key challenges in this setting is the presence of non-iid datasets across clients. Several modifications of the original FedAvg algorithm have been proposed to address this challenge. Some of these approaches focus on finding better solutions to the optimization problem to prevent the divergence of the global solution~\citep{fedprox, scaffold, fedpd, fedsplit, feddyn} whereas some suggest the modification of the training procedure to incorporate appropriate aggregation of the local models~\citep{fedbe, fedma, pfnm, fednova, singh2020model}. Creating local personalised models for each client to improve the overall method is also well researched~\citep{fedrep, GhoshCYR20, fed_mtl, ditto, mocha, SattlerMS21, local_adaption}. Also, augmenting the data distributions by creating additional data instances for learning~\citep{HaoELZLCC21, FLviaSyn, luo2021fear} has seen to help with the collaborative effort in federated learning.

\subsection{Self-supervised learning(SSL)} 
SSL aims at training high performing complex networks without the requirement of large labeled datasets. It does not require explicitly labeled data and target pairs but instead uses inherent structure in the data for representation learning. The two types of SSL methods include contrastive based and non-contrastive based SSL. The contrastive methods work on triplets of the form $(x, x', y)$ where $x$ and $x'$ are different views of the same instance and $y$ is a non-compatible data instance. The key idea is to train models by adjusting the parameters in such a way that the representations(embeddings) obtained for $x$ and $x'$ are nearly identical and that of $x$ and $y$ are unalike. Since obtaining these triplets from the dataset does not necessarily require presence of the labels, for example for images, $x$ and $x'$ could be different augmentations of the same image and $y$ could be an image of a different object, these methods could learn from large volume of data. Recent works in this field suggest many different ways of generating the data triplets for training~\citep{simclr, moco, instance_contrastive_learning, clusterfit}. Non-contrastive methods, on the other hand, work without generating explicit triplets. One form of the non-contrastive methods use clustering based methods to form a group of objects and use the cluster assignments as pseudo-labels for training the model~\citep{deepcluster, swav}. Another form of non-contrastive methods use joint embedding architectures like Siamese network but keep one of the architectures to provide target embeddings for training~\citep{chen2020simsiam, byol, mocov2}.

\subsection{Federated Self-supervised Learning} 
The interest in self-supervised federated learning is relatively new. Although MOON~\citep{moon} proposed using a contrastive loss in supervised federated learning, it was FedU~\citep{FedU} that proposed a way to leverage unlabeled data in a decentralised setting. Subsequently, two parallel works~\citep{divergenceaware}, ~\citep{ssfl} proposed frameworks for training self-supervised models in a federated setting and compared different self-supervised models. In all of these approaches though, the clients train local models of identical architectures. This common architecture is communicated by the server at the beginning of the training. In each round, the clients perform update steps on the local models and send the local models to the central server at the end of the round. The server appropriately aggregates the client models to generate the global model for that round and communicates the global model parameters back to each client. The prominent idea in these latest developments is the divergence aware updates of the local model to the global model wherein, if the local models are very different from the global model, the local models are not updated with the global model, which can be seen as an initial step towards creating more personalised models for each client.

\section{Methodology} \label{sec:methodology}
In order to enable federated learning for heterogeneous clients in an unsupervised way, we propose a new framework using self-supervised federated learning and name it Hetero-SSFL. This framework is unique as it allows heterogeneous clients to participate in the training process and also exhibits high empirical performance. The convergence guarantee of the algorithm for non-convex loss functions is also shown. In this section we first formally describe the problem setting and then elaborate on our proposed solution.

\subsection{Problem Definition}
A federated learning setting consists of a set of clients ${k \in |N|}$, with each client $k$ having local data instances $\Xx_k$ of size $n_k$ drawn from the local data distribution $\D_k$. The distributions $\D_k$ of different clients are more likely to be non-identical, for example, each client may have access to objects of only a few categories or classes different from other clients' object categories. The local objective at $k^{th}$ client is to solve for $\min_{\W_k} \mathcal{L}_k = \ell({\W_k})$ where $\ell(.)$ is some loss function. After the local optimization, the federation part of the learning allows the server to access all the clients' learnings in the form of the learned set of weights ${\W_1,\W_2,\dots,\W_N}$ and utilise it for achieving global objective. In traditional FL methods, it is assumed that all $\W_k$'s are identical in shape and in each round the server combines the local models to learn a global model $\hat{\W}$ of the form $ \hat{\W} =  g(\W_1,\W_2,\dots,\W_N)$, where $g$ is an appropriate aggregation function, and sends $\hat{\W}$ back to the clients for further updates. The novelty of our solution is in the fact that our solution does not require the local models (and thus the local weights) to be of identical size. The way we utilise the global knowledge, ${\W_1,\W_2,\dots,\W_N}$, to achieve collaboration for each client's benefit, is also different.

\subsection{Hetero-SSFL Framework}
\begin{figure}
\begin{center}
    \includegraphics[height=2.7in,width=4in]{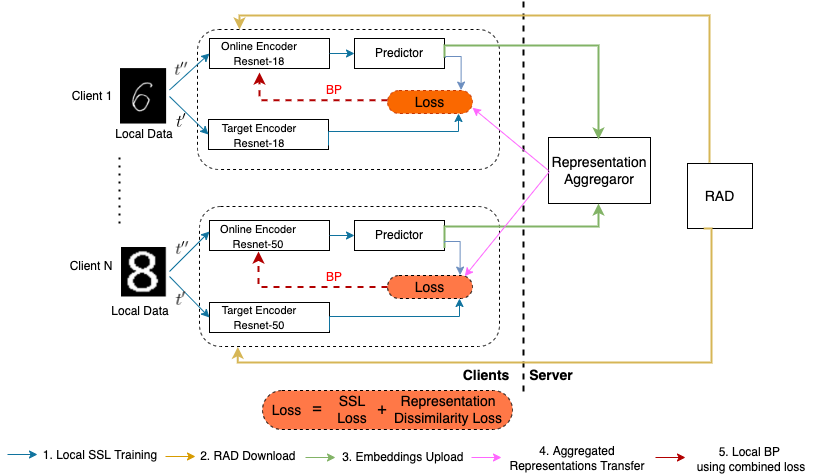}
\end{center}
  \caption{Overview of the proposed framework: Hetero-SSFL. In each training round, the server sends the RAD and the aggregated representation matrix to all clients, each client trains local self-supervised models by optimising the combined loss, the representations obtained using local model on RAD are sent back to the server which aggregates the representations and sends it back to all clients for next training round. Note: the local models are named Resnet-18 and Resnet-50 as examples to depict heterogeneous client models.}
  \label{fig:hetero_ssfl_schematic}
\end{figure}

We introduce the Hetero-SSFL framework in this section and provide details of all the components in the framework here. The end-to-end pipeline is depicted in Figure~\ref{fig:hetero_ssfl_schematic} and the algorithm is detailed in Algorithm~\ref{alg:fednn_algo_hetero}. 

\paragraph{Local Training} We assume each client contains unlabeled datasets and employ self-supervised learning on each client. While there are many ways to achieve self-supervision, in this work we focus on non-contrastive methods like BYOL~\citep{byol} and SimSiam~\citep{chen2020simsiam} which involve two embedding architectures like Siamese networks for achieving self-supervision. The two parallel networks are called the \textit{online} network and \textit{target} network respectively. The target network's task is to provide targets for regression to the online network whereas the online network is trained to learn representations. The target network consists of an encoder and the online network is composed of an encoder and a prediction layer on top of it, whose role is to learn better representations under self-supervision loss. The weights of the target network are just a moving average of the weights of the online network and the target network can thus be thought of as a stable version of the online network. In federated self-supervised learning, the local set of parameters for each client $k$ include $\W_k = \{W_k^{o}, W_k^{p}, W_k^{t}\}$ where $W_k^{o}$ parameterises the representation learning component of the online network with $W_k^{p}$ being the prediction layer and $W_k^{t}$ is the target network.
The local loss at each client $\ell(.)$ is similar to self-supervised loss used in central self-supervised learning and is given by -
$$ \ell(\W_k) = || \F(v'; [W_k^{o}, W_k^{p}]) - \F'(v''; W_k^{t}) ||^2,$$
 i.e., it is the mean squared error between the outputs of the online network and the target network on different views(augmentations), $v'$ and $v''$, of the same image $v$. $\F(,;.)$ and $\F'(,;.)$ denote the online and target network functions respectively. 

In our framework, the learning on each client is also guided by peers to bring about collaboration. We use this guidance in the form of representations that each client locally learns for a common set of instances. The intuition is that the clients can work together to uncover the hidden representation structure in the data in an unsupervised way. Specifically, we modify the local loss function at each client to contain a proximal term that measures the distance between the local representations and the representations obtained on all other clients. The local loss function at each client thus becomes -
$$ \min_{\W_k} \mathcal{L}_k = \ell(\W_k) + \hspace{0.03in} \mu \text{d} \bigg( \Phi_i(\textbf{X}; \W_i^{o}), \Bar{\Phi}(\textbf{X} ; (t-1)) \bigg)$$
where d$(.,.)$ is a suitable distance metric, and $\Bar{\Phi}$ denotes the aggregated representations of other clients
$$ \Bar{\Phi}(\textbf{X} ; (t)) =  \sum_{j=1}^{N} w_j \Phi_{j}(\textbf{X}; \W_j^{o}(t)).
$$
Here, $w_j$ is the weight given to the $j^{th}$ client and could be kept higher for clients with larger resources or learnt in the training pipeline. The representations $\Phi(\textbf{X};.)$ are obtained from the output of the client's online network under parameters $\W^o = [W^{o}, W^{p}]$ at a layer suitable for defining the representation space, and this could potentially be different from $\F$ which denotes the entire online network. In our framework, we use the output of the predictor layer for an instance $x$ as $\Phi(x;.)$.

\paragraph{Communication with the Server} Different from traditional FL where each client sends the local models to server and the server creates a global model by performing element-wise aggregation on the local model weights, we gather representations at the server. We assume that the server has access to an unlabeled dataset(which could be easily obtained in practice from open sources) and call it \emph{Representation Alignment Dataset}(RAD). In each global training round $t$, server sends the RAD to all the clients. All the clients use the RAD, $\textbf{X}$, in the loss function as described above while simultaneously training local models. After the clients finish local epochs, the final representations obtained for RAD $\X$ on each client, $\Phi_k(\textbf{X}; \W_k^{o})$, are sent to the server. The server then aggregates the local representations from all clients to form the global representation matrix $\Bar{\Phi}(\textbf{X} ; (t))$ which is then again sent to the clients for training in the next round.

\subsection{The Proximal Term} 
In this sub-section we describe the second part of the loss function and provide details on the proximal term and the distance function. 

The role of the proximal term is to compare the representations obtained from various neural networks. This requires the distance function to be able to meaningfully capture the similarity in the high dimensional representational space. Moreover, the distance function should be such that it is sensitive to the changes that affect functional behavior and does not change as much with the inconsequential modifications like change in random initializations. The types of distance functions used in the literature for comparing neural network representations include Canonical Correlation Analysis(CCA), Centered Kernel ALignment(CKA) and Procustres distance based measures~\citep{ding2021grounding, cka}. While the Procustres distance based metrics cannot compare the representations obtained in spaces of different dimensions, the CCA based metrics are not suitable for training through backpropagation. In contrast, properties of the CKA metric like ability to learn similarities between layers of different widths, invariance to random initializations, invertible linear transformations and orthogonal transformations lend it useful for use in the proximal term.

CKA measures the distance between the objects in different representation spaces by creating similarity(kernel) matrices in the individual space and then comparing these similarity matrices. The idea is to use the representational similarity matrices to characterize the representation space. CKA takes the activation matrices obtained from the network as inputs and gives a similarity score between 0 and 1 by comparing the similarities between all objects. Specifically, if $A_i \in \mathbb{R}^{L \times d_i}$ and $A_j \in \mathbb{R}^{L \times d_j}$ are the activation(representation) matrices for clients $i$ and $j$ for the RAD \textbf{X} of size $L$, the distance between $A_i$ and $A_j$ using Linear-CKA is computed as - 
\begin{equation}
\begin{multlined}
\label{eqn:lin_cka}
\text{Linear CKA}(A_i A_i^T, A_j A_j^T) = \text{Linear CKA}(K_i,K_j) = \dfrac{||A_j^{T}A_i||^2_F}{||A_i^{T}A_i||_F ||A_j^{T}A_j||_F},
\end{multlined}
\end{equation}
where $K_i$ and $K_j$ are kernel matrices for any choice of kernel $\mathcal{K}$ such that we have $K_i(p,q) = \mathcal{K}(A_i(p,:), A_i(q,:))$. CKA also allows choice of other kernels like RBF-kernel, polynomial kernel etc.
but we observed that the results with Linear-CKA are as good as those with RBF kernel as also reported in~\citep{cka}. Thus we focus only on Linear-CKA here. 

Finally, the local loss at each client becomes
\begin{equation}\label{eqn:hetero}
\begin{multlined}
\min_{\W_k} \mathcal{L}_k = 
\ell({\W_k}) + \mu \hspace{0.03in} \text{Linear-CKA} \bigg( K_k, \Bar{K}(t-1) \bigg),
\end{multlined}
\end{equation}
with $ \Bar{K}(t-1)  =  \sum_{j=1}^{N} w_j K_j(t-1). $ When $\mu = 0$, each client is training its own local models without any peer-supervision and when $\mu \rightarrow \infty$ the local models try to converge on the RAD representations without caring about the SSL loss. In such cases, if the size of RAD is increased arbitrarily and models have similar architectures, the local models become identical.

\begin{algorithm}[tb]
   \caption{Hetero-SSFL Algorithm for Heterogeneous clients}
   \label{alg:fednn_algo_hetero}
   \begin{algorithmic}
   \STATE {\bfseries Input:} number of clients $N$, number of global communication rounds $T$, 
   number of local epochs $E$, parameter $\mu$, weight vector for clients $[w_1,w_2,\dots w_n]$ 
   \STATE {\bfseries Output:} Final set of personalised models {$\mathcal{W}_1(T), \mathcal{W}_2(T) \dots \mathcal{W}_n(T)$} \\
   
   \STATE {\bfseries At Server - }
   \STATE Initialize {$\mathcal{W}_1(0), \mathcal{W}_2(0) \dots \mathcal{W}_N(0)$}
   \FOR{$t=1$ {\bfseries to} $T$}
    \STATE RAD = \textbf{X}
    \FOR{each client $j$}
    \STATE $A_j(\textbf{X}) = \Phi_j(\textbf{X};\mathcal{W}_j(t-1))$
    \STATE $K_j = \mathcal{K}(A_j(\textbf{X}),A_j(\textbf{X}))$
    \ENDFOR
    \STATE $\Bar{K}(t-1)  =  \sum_{j=1}^{N} w_j K_j$
    \STATE Select a subset of clients $\mathcal{N}_t$  
    \FOR{each selected client $i \in \mathcal{N}_t$}
    \STATE $\mathcal{W}_i(t) =$ \textbf{LocalTraining}$( \mathcal{W}_i(t-1), \Bar{K}(t-1), \textbf{X}, \mu)$
   \ENDFOR
   \ENDFOR
   \STATE Return {$\mathcal{W}_1(T), \mathcal{W}_2(T) \dots \mathcal{W}_n(T)$}
   \STATE {\bfseries LocalTraining}$(\mathcal{W}_i(t-1), \Bar{K}(t-1), \textbf{X}, \mu)$
   \STATE Initialize $\mathcal{W}_i(t)$ with $(\mathcal{W}_i(t-1))$
   \FOR{each local epoch}
   \STATE  $A_i(\textbf{X}) = \Phi_i(\textbf{X};\mathcal{W}_i(t))$
   \STATE  $K_i = \mathcal{K}(A_i(\textbf{X}),A_i(\textbf{X}))$
   \STATE Update $\mathcal{W}_i(t)$ using SGD for loss in equation ~\eqref{eqn:hetero}
   \ENDFOR
   \STATE Return $\mathcal{W}_i(t)$  
\end{algorithmic}
\end{algorithm}

\section{Convergence Analysis} \label{sec:analysis}
In this section we provide insights on convergence of the proposed framework. The convergence guarantee is shown to hold under mild assumptions which are commonly made in the literature~\citep{fednova, fedproto}.

\begin{assumption}[Lipschitz Smoothness]\label{loss_assum} The local loss function on every client $k$, $\loss_k$, is assumed to be $L_1$-Lipschitz smooth, which implies:
\begin{equation}
\loss_k(a) - \loss_k(b) \leq \nabla \loss_k(b)^{\top} (\W_k(a) - \W_k(b)) + \frac{L_1}{2}||\W_k(a) - \W_k(b)||_2^2 , \hspace{0.1in} \forall \hspace{0.05in} a, b.
\end{equation}
\end{assumption}

\begin{assumption}[Stochastic Gradient]\label{gd_assum} The local stochastic gradient for each client $k$, at any time $t$, $g_{k,t}$ is an unbiased estimator of the gradient, has bounded variance and the expectation of its norm is bounded above. That is, we have
$$\E[g_{k,t}] = \nabla \loss_k \hspace{0.1in} \text{and} \hspace{0.1in} \text{Var}(g_{k,t}) \leq \sigma^2,$$
$$\E[||g_{k,t}||_2] \leq P.$$
\end{assumption}

\begin{assumption}[Representation Norm]\label{rep_assum} The norm of the representations obtained for any instance on any client is bounded above by $R$. That is:
$$ ||\Phi_k(x;\W_k)||_2 \leq R.$$
\end{assumption}

\begin{assumption}[Lipschitz Continuity]\label{phi_assum} The embedding function $\Phi(.;.)$ for all clients is $L_2$-Lipschitz continuous, which implies:
$$||\Phi_k(.;\W_a) - \Phi_k(.;\W_b)||_2 \leq L_2 ||\W_a - \W_b||_2. $$
\end{assumption}
We start with the following lemma on the reduction of the local loss after $E$ local epochs.
\begin{lemma}\label{local_epochs}
(Local Epochs) In each communication round, the local loss $\loss$ for each client reduces after $E$ local epochs and is bounded as below under the Assumption~\ref{gd_assum}.
\begin{equation}\label{bound1}
\E[\loss_{E}] \leq  \loss_{0} - (\eta - \dfrac{L_1 \eta^2}{2})\sum_{i=0}^{E-1}||\nabla \mathcal{L}_{i} ||^2 + \dfrac{L_1 E \eta^2}{2} \sigma^2.
\end{equation}
where $\loss_{0}$ and $\loss_{E}$ denote the loss before and after $E$ local epochs respectively and $\eta$ is learning rate.
\end{lemma}
Proof of Lemma~\ref{local_epochs} is in Appendix. Lemma~\ref{local_epochs} shows the bound on the client's local loss after the completion of local epochs under one global communication round.

After the local training, the representations from all clients are sent to the server. Our next lemma~\ref{rep_update} shows a bound on the expected loss after every global representation update at the server. 

\begin{lemma}\label{rep_update}
(Representation Update) In each global communication round, for each client, after the $E$ local updates, the global representation matrix is updated at the server and the loss function for any client $k$ gets modified to $\loss_{E'}$ from $\loss_{E}$ and could be bounded as -
\begin{equation}\label{bound2}
\E[\loss_{E'}] \leq \E[\loss_{E}] + 2 \mu \eta \text{L}_2 P R^3 L^2.
\end{equation}
\end{lemma}
Proof of Lemma~\ref{rep_update} is in Appendix.

Equipped with the results from Lemmas~\ref{local_epochs} and~\ref{rep_update}, we are now ready to state our main result which provides a total deviation bound for an entire training round. Theorem~\ref{theorem:convergence} shows the divergence in loss for any client after completion of one round. We can guarantee the convergence by appropriately choosing $\mu$ and $\eta$ which leads to a certain expected decrease in the loss function. The result holds for convex as well as non-convex loss functions.

\begin{theorem}[Convergence]
\label{theorem:convergence} After one global round, the loss function of any client decreases and is bounded as shown below:
\begin{equation}
\E[\loss_{E'}] \leq \loss_0 - (\eta - \dfrac{L_1 \eta^2}{2})\sum_{i=0}^{E-1}||\nabla \mathcal{L}_{i} ||^2 + \dfrac{L_1 E \eta^2}{2} \sigma^2 + 2 \mu \eta \text{L}_2 P R^3 L^2.  
\end{equation}
Thus, if we choose $\eta$ and $\mu$ such that
\begin{align*}
\eta < \dfrac{2(\sum_{i=0}^{E-1}||\nabla \mathcal{L}_{i} ||^2 - 2\mu \text{L}_2 P R^3 L^2)}{L_1 (\sum_{i=0}^{E-1} ||\nabla \mathcal{L}_{i} ||^2 + E \sigma^2)}, \quad \mu < \dfrac{\sum_{i=0}^{E-1}||\nabla \mathcal{L}_{i} ||^2 }{2 \text{L}_2 P R^3 L^2},
\end{align*}
then the convergence of Algorithm~\ref{alg:fednn_algo_hetero} is guaranteed.
\end{theorem}
The proof of Theorem~\ref{theorem:convergence} follows directly from the results of Lemmas~\ref{local_epochs} and~\ref{rep_update}. 

\section{Experiments} \label{sec:experiments}
In this section we describe the experimental setup and present results for our method alongside the popular baselines. We simulate system heterogeneity by randomly choosing different architectures for local client models. Within system heterogeneity we evaluate our model under varying statistical settings by manipulating the data distributions across clients to be IID or non-IID. We also demonstrate the performance of our framework with increase in number of clients. 

\subsection{Experimental Details}
\paragraph{Datasets} We test our framework on image classification task and use three datasets to compare against the baselines: CIFAR-10, CIFAR-100 and Tiny-Imagenet as suggested by several works in FL and the popular federated learning benchmark LEAF \citep{caldas2019leaf}. CIFAR-10 and CIFAR-100 contain 50,000 train and 10,000 test colored images for 10 classes and 100 classes respectively and Tiny-Imagenet has 100,000 images for 200 classes. 

\paragraph{Implementation Details} For the FL simulations, we explore two types of settings, non-IID setting and IID setting. In the IID-setting, we assume each client to have access to the data of all classes and in the non-IID setting we assume each client has access to data of only a few disjoint classes. For example, for CIFAR-10 client 1 might have access to objects of classes $\{1,2\}$ versus client 2 with access to $\{3,4\}$ and likewise. The IID setting is created by dividing all the instances between the clients. For the non-IID setting, for each client a fraction of classes is sampled and then instances are divided amongst the clients containing specific classes. In the cases when number of clients, $N$, is less than total number of classes $K$, each client is assumed to have access to all instances corresponding to $\frac{K}{N}$ classes. The choice
of encoders at each client is data and application dependent. Since we are dealing with images, to create heterogeneous clients we use Resnet-18 and Resnet-34 models as encoders for our method. Since all the baselines work in homogeneous client settings and hence have to use the same model, we use Resnet-34  at each client in the baselines for comparison. The total number of global communication rounds is kept to 200 for all methods with each client performing $E = 5$ local epochs in each round. We use SGD with momentum $= 0.9$, batch size $= 200$ and learning rate $= .032$ for training all models. The weight vector used in aggregating the representations $\textbf{w}$ is set to be uniform to $\frac{1}{n}$. All these models are trained on a machine with 4 GeForce RTX 3090 GPUs with each GPU having about 24gb of memory.

\paragraph{Baselines} We compare our method against two very recently proposed federated self-supervised baselines: FedU and FedEMA, which reflect the current state-of-the-art. Since the source code for these baselines is not publicly available, (though they have been promised in the future as per our personal communications), we perform these comparisons using our implementation and with the best parameters reported in these papers. We also create additional baselines by combining central SSL algorithms like BYOL and SimSiam with the federated learning procedure of FedAvg to make FedBYOL and FedSimSiam. The performance of supervised FedAvg and centralised BYOL is also reported for comparisons as theoretical baselines. All of these methods however work in homogeneous settings, hence we compare the homogeneous setting of baselines with Resnet-34 architecture with our heterogeneous setting having a mix of Resnet-34 and Resnet-18 architectures. One additional baseline is the SSFL algorithm, but due to the unavailability of the source code and lack of precise implementation details in the paper, we could not compare against it. 

\paragraph{Evaluation} We follow the linear evaluation protocol as mentioned in~\citep{FedU} to test the models. Under this protocol, the local encoders are trained in self-supervised way without any labeled data and then a linear classifier is trained on top of the encoder after freezing the encoder parameters. We use a linear fully connected layer with 1000 neurons as the classifier and train it for 100 epochs with Adam optimizer, batch size = 512 and learning rate = 0.003. The reported results are the accuracy of the linear classifier on a separate test dataset.

\paragraph{Other parameters} The other parameters important for our approach are $\mu$ which controls the weight of the proximal term in the loss function, and the size of the RAD dataset. We tune the hyperparameter $\mu$ using a validation set and found the best values of $\mu$ to be $0.5$ for the CIFAR datasets and $1$ for the Tiny-Imagenet dataset. For the RAD size we observe that increasing the size of RAD increases performance but has an effect on the speed of simulation so we fix that size to be 5000 in our experiments. On studying the effect of varying number of local epochs we observed that unlike the FedAvg based algorithms our method is stable with increasing local epochs. 

\begin{table}
  \caption{Test Accuracy comparison of our method with baselines under the linear evaluation protocol in non-IID settings.}
  \label{baselines_noniid}
  \centering
  \begin{tabular}{lccr}
    \toprule
    \cmidrule(r){1-2}
    Method     &  CIFAR-10 & CIFAR-100 & Tiny-Imagenet \\
    \midrule
    FedBYoL     &  78.1 $\pm$ 1.1   &  58.1 $\pm$ 0.3   &  31.8 $\pm$ 2.0  \\
    FedSimSiam  &  76.7 $\pm$ 1.5   &  53.2 $\pm$ 0.16  &  29.1   $\pm$ 1.2 \\
    FedU        &  79.6 $\pm$ 0.5   &  58.9 $\pm$ 0.12  &  58.23  $\pm$ 2.3 \\
    FedEMA      &  81.2 $\pm$ 1.6   &  61.8 $\pm$ 0.31  &  58.2  $\pm$ 3.1  \\
    Hetero-SSFL &  \textbf{90.29 $\pm$ 0.3}  &  \textbf{67.7 $\pm$ 0.21}  &  \textbf{61.5 $\pm$ 1.8}  \\
    \midrule
    Supervised FedAvg &  68.5 $\pm$ 1.7 & 44.29 $\pm$ 1.34  &  27.2 $\pm$ 3.2 \\
    Central-BYOL (IID upper bound) & 94.3 $\pm$ 0.2  &  74.2 $\pm$ 0.67 & 78.7 $\pm$ 1.3 \\
    \bottomrule
  \end{tabular}
\end{table}

\begin{table}[!htb]
  \caption{Test Accuracy comparison of our method with baselines under the linear evaluation protocol in IID settings(with equal number of classes on all clients).}
  \label{baselines_iid}
  \centering
  \begin{tabular}{lccccr}
    \toprule
    \cmidrule(r){1-2}
    Dataset  &  FedBYoL & FedSimSiam & FedU & FedEMA & Hetero-SSFL \\
    \midrule
    CIFAR-10   &  82.24 $\pm$ 0.8   &  80.4 $\pm$ 1.2  &  81.6 $\pm$ 1.8   &  85.9 $\pm$ 0.3  &  \textbf{88.5 $\pm$ 1.3}  \\
    CIFAR-100  &  23.8 $\pm$ 0.14   &  22.19 $\pm$ 1.8 &  20.4 $\pm$ 0.91  &  25.7  $\pm$ 0.4  & \textbf{34.8 $\pm$ 0.7}  \\
    \bottomrule
  \end{tabular}
\end{table}

\subsection{Results}
We show the performance of our method in comparison with baselines in non-IID settings in Table~\ref{baselines_noniid} and that for IID settings in Table~\ref{baselines_iid}. We observe that our method outperforms the baselines by a significant margin and is slightly worse than the central BYOL which implies that we are able to achieve successful collaboration amongst clients. We also report the performance of our method and high-performing baselines with increase in scale in Table~\ref{scalable_c100}. As the number of clients increase, the assumption on high availability of clients is relaxed and in every round only a fraction of clients are sampled to perform local training, total 20 clients with 5 clients selected per round is denoted as 20clients(5) in Table~\ref{scalable_c100}. For additional insights, we show the average change in test accuracy of clients with different architectures when performing standalone training versus collaborative learning in the heterogeneous setting.

\begin{table}[!htb]
  \caption{Test Accuracy of various federated self-supervised methods with change in scale on CIFAR-100 dataset.}
  \label{scalable_c100}
  \centering
  \begin{tabular}{lcccr}
    \toprule
    \cmidrule(r){1-2}
    Dataset  &  5clients(5) &  20clients(5)  &  100clients(10) \\
    \midrule
    FedU        & 58.9 $\pm$ 0.12   &  48.29 $\pm$ 1.8 & 41.1 $\pm$ 2.3\\
    FedEMA      & 61.8 $\pm$ 0.31   &  52.5 $\pm$ 0.8  & 42.0 $\pm$ 2.7\\
    Hetero-SSFL & 67.7 $\pm$ 0.21  &  57.5  $\pm$ 0.9  & 43.3 $\pm$ 1.9 \\
    \bottomrule
  \end{tabular}
\end{table}

\begin{table}[!htb]
  \caption{Average effect of collaboration on different clients. Column 2 depicts the test accuracy of clients with local models and column 3 the accuracy under Hetero-SSFL.}
  \label{collab}
  \centering
  \begin{tabular}{lccr}
    \toprule
    \cmidrule(r){1-2}
    Client Model Architecture  &  Local SSL &  Hetero-SSFL \\
    \midrule
    Resnet-18    & 71.9   &  85.4 \\
    Resnet-34    & 77.8   &  93.7 \\
    \bottomrule
  \end{tabular}
\end{table}

\section{Conclusion} \label{sec:discussion}
In this work we study the problem of self-supervised federated learning with heterogeneous clients and propose a novel framework to enable collaboration in such settings. The framework allows clients with  varying data/compute resources to partake in the collaborative training procedure. This development enables several practical settings to still gainfully employ self-supervised FL even though the  different clients, for example different edge-devices or distinct organisations, might not have similar compute and data resources. The high performance of the framework across multiple datasets indicates its promise in achieving collaboration while the theoretical results guarantee the convergence of the algorithm for fairly general loss functions. In future work we plan to investigate the use of attention based mechanisms for adaptive weighting of different clients to generate aggregate representations based on each client's relative importance. Also, rather than assuming that the server has access to an unlabeled dataset and transmitting RAD from the server to clients, we can create a common generative model in a federated way like federated GANs to generate data instances required for RAD.

\newpage
\bibliography{refs}

\begin{thebibliography}{51}
\providecommand{\natexlab}[1]{#1}
\providecommand{\url}[1]{\texttt{#1}}
\expandafter\ifx\csname urlstyle\endcsname\relax
  \providecommand{\doi}[1]{doi: #1}\else
  \providecommand{\doi}{doi: \begingroup \urlstyle{rm}\Url}\fi

\bibitem[Acar et~al.(2021)Acar, Zhao, Matas, Mattina, Whatmough, and
  Saligrama]{feddyn}
Durmus Alp~Emre Acar, Yue Zhao, Ramon Matas, Matthew Mattina, Paul Whatmough,
  and Venkatesh Saligrama.
\newblock Federated learning based on dynamic regularization.
\newblock In \emph{International Conference on Learning Representations}, 2021.

\bibitem[Aggarwal and Yu(2008)]{aggarwal2008general}
Charu~C Aggarwal and Philip~S Yu.
\newblock A general survey of privacy-preserving data mining models and
  algorithms.
\newblock In \emph{Privacy-preserving data mining}, pages 11--52. Springer,
  2008.

\bibitem[Caldas et~al.(2019)Caldas, Duddu, Wu, Li, Konečný, McMahan, Smith,
  and Talwalkar]{caldas2019leaf}
Sebastian Caldas, Sai Meher~Karthik Duddu, Peter Wu, Tian Li, Jakub Konečný,
  H.~Brendan McMahan, Virginia Smith, and Ameet Talwalkar.
\newblock Leaf: A benchmark for federated settings, 2019.

\bibitem[Caron et~al.(2018)Caron, Bojanowski, Joulin, and Douze]{deepcluster}
Mathilde Caron, Piotr Bojanowski, Armand Joulin, and Matthijs Douze.
\newblock Deep clustering for unsupervised learning of visual features.
\newblock In Vittorio Ferrari, Martial Hebert, Cristian Sminchisescu, and Yair
  Weiss, editors, \emph{Computer Vision -- ECCV 2018}, pages 139--156. Springer
  International Publishing, 2018.

\bibitem[Caron et~al.(2020)Caron, Misra, Mairal, Goyal, Bojanowski, and
  Joulin]{swav}
Mathilde Caron, Ishan Misra, Julien Mairal, Priya Goyal, Piotr Bojanowski, and
  Armand Joulin.
\newblock Unsupervised learning of visual features by contrasting cluster
  assignments.
\newblock In Hugo Larochelle, Marc'Aurelio Ranzato, Raia Hadsell,
  Maria{-}Florina Balcan, and Hsuan{-}Tien Lin, editors, \emph{Advances in
  Neural Information Processing Systems 33: Annual Conference on Neural
  Information Processing Systems 2020, NeurIPS 2020, December 6-12, 2020,
  virtual}, 2020.

\bibitem[Chen and Chao(2021)]{fedbe}
Hong-You Chen and Wei-Lun Chao.
\newblock Fed{\{}be{\}}: Making bayesian model ensemble applicable to federated
  learning.
\newblock In \emph{International Conference on Learning Representations}, 2021.

\bibitem[Chen et~al.(2020{\natexlab{a}})Chen, Kornblith, Norouzi, and
  Hinton]{simclr}
Ting Chen, Simon Kornblith, Mohammad Norouzi, and Geoffrey Hinton.
\newblock A simple framework for contrastive learning of visual
  representations.
\newblock In \emph{Proceedings of the 37th International Conference on Machine
  Learning}, volume 119 of \emph{Proceedings of Machine Learning Research},
  pages 1597--1607. PMLR, 2020{\natexlab{a}}.

\bibitem[Chen and He(2020)]{chen2020simsiam}
Xinlei Chen and Kaiming He.
\newblock Exploring simple siamese representation learning.
\newblock \emph{arXiv preprint arXiv:2011.10566}, 2020.

\bibitem[Chen et~al.(2020{\natexlab{b}})Chen, Fan, Girshick, and He]{mocov2}
Xinlei Chen, Haoqi Fan, Ross Girshick, and Kaiming He.
\newblock Improved baselines with momentum contrastive learning,
  2020{\natexlab{b}}.
\newblock cite arxiv:2003.04297Comment: Tech report, 2 pages + references.

\bibitem[Collins et~al.(2021)Collins, Hassani, Mokhtari, and
  Shakkottai]{fedrep}
Liam Collins, Hamed Hassani, Aryan Mokhtari, and Sanjay Shakkottai.
\newblock Exploiting shared representations for personalized federated
  learning.
\newblock \emph{arXiv preprint arXiv:2102.07078}, 2021.

\bibitem[Ding et~al.(2021)Ding, Denain, and Steinhardt]{ding2021grounding}
Frances Ding, Jean-Stanislas Denain, and Jacob Steinhardt.
\newblock Grounding representation similarity through statistical testing.
\newblock In A.~Beygelzimer, Y.~Dauphin, P.~Liang, and J.~Wortman Vaughan,
  editors, \emph{Advances in Neural Information Processing Systems}, 2021.

\bibitem[Fallah et~al.(2020)Fallah, Mokhtari, and Ozdaglar]{perFL}
Alireza Fallah, Aryan Mokhtari, and Asuman~E. Ozdaglar.
\newblock Personalized federated learning: {A} meta-learning approach.
\newblock \emph{CoRR}, abs/2002.07948, 2020.

\bibitem[Gan et~al.(2017)Gan, Lin, Chao, and Zhan]{gali17}
Wensheng Gan, Chun-Wei Lin, Han-Chieh Chao, and Justin Zhan.
\newblock Data mining in distributed environment: a survey.
\newblock \emph{Wiley Interdisciplinary Reviews: Data Mining and Knowledge
  Discovery}, 7:\penalty0 e1216, 07 2017.
\newblock \doi{10.1002/widm.1216}.

\bibitem[Ghosh et~al.(2020)Ghosh, Chung, Yin, and Ramchandran]{GhoshCYR20}
Avishek Ghosh, Jichan Chung, Dong Yin, and Kannan Ramchandran.
\newblock An efficient framework for clustered federated learning.
\newblock In Hugo Larochelle, Marc'Aurelio Ranzato, Raia Hadsell,
  Maria{-}Florina Balcan, and Hsuan{-}Tien Lin, editors, \emph{Advances in
  Neural Information Processing Systems 33: Annual Conference on Neural
  Information Processing Systems 2020, NeurIPS 2020, December 6-12, 2020,
  virtual}, 2020.

\bibitem[Goetz and Tewari(2020)]{FLviaSyn}
Jack Goetz and Ambuj Tewari.
\newblock Federated learning via synthetic data.
\newblock \emph{CoRR}, abs/2008.04489, 2020.

\bibitem[Grill et~al.(2020)Grill, Strub, Altch\'{e}, Tallec, Richemond,
  Buchatskaya, Doersch, Avila~Pires, Guo, Gheshlaghi~Azar, Piot, kavukcuoglu,
  Munos, and Valko]{byol}
Jean-Bastien Grill, Florian Strub, Florent Altch\'{e}, Corentin Tallec, Pierre
  Richemond, Elena Buchatskaya, Carl Doersch, Bernardo Avila~Pires, Zhaohan
  Guo, Mohammad Gheshlaghi~Azar, Bilal Piot, koray kavukcuoglu, Remi Munos, and
  Michal Valko.
\newblock Bootstrap your own latent - a new approach to self-supervised
  learning.
\newblock In H.~Larochelle, M.~Ranzato, R.~Hadsell, M.F. Balcan, and H.~Lin,
  editors, \emph{Advances in Neural Information Processing Systems}, volume~33,
  pages 21271--21284. Curran Associates, Inc., 2020.

\bibitem[Hao et~al.(2021)Hao, El{-}Khamy, Lee, Zhang, Liang, Chen, and
  Carin]{HaoELZLCC21}
Weituo Hao, Mostafa El{-}Khamy, Jungwon Lee, Jianyi Zhang, Kevin~J. Liang,
  Changyou Chen, and Lawrence Carin.
\newblock Towards fair federated learning with zero-shot data augmentation.
\newblock In \emph{{IEEE} Conference on Computer Vision and Pattern Recognition
  Workshops, {CVPR} Workshops 2021, virtual, June 19-25, 2021}, pages
  3310--3319. Computer Vision Foundation / {IEEE}, 2021.

\bibitem[He et~al.(2022)He, Yang, Mushtaq, Lee, Soltanolkotabi, and
  Avestimehr]{ssfl}
Chaoyang He, Zhengyu Yang, Erum Mushtaq, Sunwoo Lee, Mahdi Soltanolkotabi, and
  Salman Avestimehr.
\newblock {SSFL}: Tackling label deficiency in federated learning via
  personalized self-supervision, 2022.

\bibitem[He et~al.(2020)He, Fan, Wu, Xie, and Girshick]{moco}
Kaiming He, Haoqi Fan, Yuxin Wu, Saining Xie, and Ross Girshick.
\newblock Momentum contrast for unsupervised visual representation learning.
\newblock In \emph{2020 IEEE/CVF Conference on Computer Vision and Pattern
  Recognition (CVPR)}, pages 9726--9735, 2020.
\newblock \doi{10.1109/CVPR42600.2020.00975}.

\bibitem[Jeong et~al.(2021)Jeong, Yoon, Yang, and Hwang]{semi-federated}
Wonyong Jeong, Jaehong Yoon, Eunho Yang, and Sung~Ju Hwang.
\newblock Federated semi-supervised learning with inter-client consistency \&
  disjoint learning.
\newblock In \emph{International Conference on Learning Representations}, 2021.

\bibitem[Jiang et~al.(2020)Jiang, Kone{\v{c}}n{\'y}, Rush, and
  Kannan]{jiang2020improving}
Yihan Jiang, Jakub Kone{\v{c}}n{\'y}, Keith Rush, and Sreeram Kannan.
\newblock Improving federated learning personalization via model agnostic meta
  learning, 2020.

\bibitem[Kargupta and Park(2000)]{kapa00}
H.~Kargupta and B.-H. Park.
\newblock Collective data mining: A new perspective toward distributed data
  mining.
\newblock In H.~Kargupta and P.~Chan, editors, \emph{Advances in Distributed
  and Parallel Knowledge Discovery}. AAAI/MIT Press, 2000.

\bibitem[Kargupta et~al.(2001)Kargupta, Huang, Krishnamoorthy, and
  Johnson]{kahu01}
H.~Kargupta, W.~Huang, Krishnamoorthy, and E.~Johnson.
\newblock Distributed clustering using collective principal component analysis.
\newblock \emph{Knowledge and Information Systems Journal Special Issue on
  Distributed and Parallel Knowledge Discovery}, 3:\penalty0 422--448, 2001.

\bibitem[Karimireddy et~al.(2020)Karimireddy, Kale, Mohri, Reddi, Stich, and
  Suresh]{scaffold}
Sai~Praneeth Karimireddy, Satyen Kale, Mehryar Mohri, Sashank Reddi, Sebastian
  Stich, and Ananda~Theertha Suresh.
\newblock {SCAFFOLD}: Stochastic controlled averaging for federated learning.
\newblock In Hal~Daumé III and Aarti Singh, editors, \emph{Proceedings of the
  37th International Conference on Machine Learning}, volume 119 of
  \emph{Proceedings of Machine Learning Research}, pages 5132--5143. PMLR,
  13--18 Jul 2020.

\bibitem[Kornblith et~al.(2019)Kornblith, Norouzi, Lee, and Hinton]{cka}
Simon Kornblith, Mohammad Norouzi, Honglak Lee, and Geoffrey Hinton.
\newblock Similarity of neural network representations revisited.
\newblock In Kamalika Chaudhuri and Ruslan Salakhutdinov, editors,
  \emph{Proceedings of the 36th International Conference on Machine Learning},
  volume~97 of \emph{Proceedings of Machine Learning Research}, pages
  3519--3529. PMLR, 2019.

\bibitem[Li et~al.(2021{\natexlab{a}})Li, He, and Song]{moon}
Qinbin Li, Bingsheng He, and Dawn Song.
\newblock Model-contrastive federated learning.
\newblock In \emph{Proceedings of the IEEE/CVF Conference on Computer Vision
  and Pattern Recognition}, 2021{\natexlab{a}}.

\bibitem[Li et~al.(2020)Li, Sahu, Zaheer, Sanjabi, Talwalkar, and
  Smith]{fedprox}
Tian Li, Anit~Kumar Sahu, Manzil Zaheer, Maziar Sanjabi, Ameet Talwalkar, and
  Virginia Smith.
\newblock Federated optimization in heterogeneous networks.
\newblock In Inderjit~S. Dhillon, Dimitris~S. Papailiopoulos, and Vivienne Sze,
  editors, \emph{Proceedings of Machine Learning and Systems 2020, MLSys 2020,
  Austin, TX, USA, March 2-4, 2020}. mlsys.org, 2020.

\bibitem[Li et~al.(2021{\natexlab{b}})Li, Hu, Beirami, and Smith]{ditto}
Tian Li, Shengyuan Hu, Ahmad Beirami, and Virginia Smith.
\newblock Ditto: Fair and robust federated learning through personalization.
\newblock In \emph{Proceedings of the 38th International Conference on Machine
  Learning, {ICML} 2021, 18-24 July 2021, Virtual Event}, volume 139 of
  \emph{Proceedings of Machine Learning Research}, pages 6357--6368. {PMLR},
  2021{\natexlab{b}}.

\bibitem[Lu et~al.(2022)Lu, Wang, Li, Niu, Dou, and Sugiyama]{semi-federated2}
Nan Lu, Zhao Wang, Xiaoxiao Li, Gang Niu, Qi~Dou, and Masashi Sugiyama.
\newblock Federated learning from only unlabeled data with
  class-conditional-sharing clients.
\newblock In \emph{International Conference on Learning Representations}, 2022.

\bibitem[Luo et~al.(2021)Luo, Chen, Hu, Zhang, Liang, and Feng]{luo2021fear}
Mi~Luo, Fei Chen, Dapeng Hu, Yifan Zhang, Jian Liang, and Jiashi Feng.
\newblock No fear of heterogeneity: Classifier calibration for federated
  learning with non-iid data, 2021.

\bibitem[Makhija et~al.(2022)Makhija, Han, Ho, and Ghosh]{fedhenn}
Disha Makhija, Xing Han, Nhat Ho, and Joydeep Ghosh.
\newblock Architecture agnostic federated learning for neural networks.
\newblock In \emph{ICML}, 2022.

\bibitem[McMahan et~al.(2017)McMahan, Moore, Ramage, Hampson, and
  Arcas]{pmlr-v54-mcmahan17a}
Brendan McMahan, Eider Moore, Daniel Ramage, Seth Hampson, and Blaise Aguera~y
  Arcas.
\newblock {Communication-Efficient Learning of Deep Networks from Decentralized
  Data}.
\newblock In Aarti Singh and Jerry Zhu, editors, \emph{Proceedings of the 20th
  International Conference on Artificial Intelligence and Statistics},
  volume~54 of \emph{Proceedings of Machine Learning Research}, pages
  1273--1282. PMLR, 20--22 Apr 2017.

\bibitem[Merugu and Ghosh(2005)]{megh05}
S.~Merugu and J.~Ghosh.
\newblock A distributed learning framework for heterogeneous data sources.
\newblock In \emph{Proc. KDD}, pages 208--217, 2005.

\bibitem[Merugu and Ghosh(Nov, 2003)]{megh03}
S.~Merugu and J.~Ghosh.
\newblock Privacy perserving distributed clustering using generative models.
\newblock In \emph{Proc. ICDM}, pages 211--218, Nov, 2003.

\bibitem[Pathak and Wainwright(2020)]{fedsplit}
Reese Pathak and Martin~J Wainwright.
\newblock Fedsplit: an algorithmic framework for fast federated optimization.
\newblock In H.~Larochelle, M.~Ranzato, R.~Hadsell, M.~F. Balcan, and H.~Lin,
  editors, \emph{Advances in Neural Information Processing Systems}, volume~33,
  pages 7057--7066. Curran Associates, Inc., 2020.

\bibitem[Sattler et~al.(2021)Sattler, M{\"{u}}ller, and Samek]{SattlerMS21}
Felix Sattler, Klaus{-}Robert M{\"{u}}ller, and Wojciech Samek.
\newblock Clustered federated learning: Model-agnostic distributed multitask
  optimization under privacy constraints.
\newblock \emph{{IEEE} Trans. Neural Networks Learn. Syst.}, 32\penalty0
  (8):\penalty0 3710--3722, 2021.

\bibitem[Singh and Jaggi(2020)]{singh2020model}
Sidak~Pal Singh and Martin Jaggi.
\newblock Model fusion via optimal transport.
\newblock \emph{Advances in Neural Information Processing Systems}, 33, 2020.

\bibitem[Smith et~al.(2017{\natexlab{a}})Smith, Chiang, Sanjabi, and
  Talwalkar]{fed_mtl}
Virginia Smith, Chao-Kai Chiang, Maziar Sanjabi, and Ameet~S Talwalkar.
\newblock Federated multi-task learning.
\newblock In I.~Guyon, U.~V. Luxburg, S.~Bengio, H.~Wallach, R.~Fergus,
  S.~Vishwanathan, and R.~Garnett, editors, \emph{Advances in Neural
  Information Processing Systems}, volume~30. Curran Associates, Inc.,
  2017{\natexlab{a}}.

\bibitem[Smith et~al.(2017{\natexlab{b}})Smith, Chiang, Sanjabi, and
  Talwalkar]{mocha}
Virginia Smith, Chao{-}Kai Chiang, Maziar Sanjabi, and Ameet~S. Talwalkar.
\newblock Federated multi-task learning.
\newblock In Isabelle Guyon, Ulrike von Luxburg, Samy Bengio, Hanna~M. Wallach,
  Rob Fergus, S.~V.~N. Vishwanathan, and Roman Garnett, editors, \emph{Advances
  in Neural Information Processing Systems 30: Annual Conference on Neural
  Information Processing Systems 2017, December 4-9, 2017, Long Beach, CA,
  {USA}}, pages 4424--4434, 2017{\natexlab{b}}.

\bibitem[Tan et~al.(2022)Tan, Long, Liu, Zhou, Lu, Jiang, and Zhang]{fedproto}
Yue Tan, Guodong Long, Lu~Liu, Tianyi Zhou, Qinghua Lu, Jing Jiang, and Chengqi
  Zhang.
\newblock Fedproto: Federated prototype learning across heterogeneous clients.
\newblock In \emph{AAAI Conference on Artificial Intelligence}, 2022.

\bibitem[Wang et~al.(2020{\natexlab{a}})Wang, Yurochkin, Sun, Papailiopoulos,
  and Khazaeni]{fedma}
Hongyi Wang, Mikhail Yurochkin, Yuekai Sun, Dimitris Papailiopoulos, and
  Yasaman Khazaeni.
\newblock Federated learning with matched averaging.
\newblock In \emph{International Conference on Learning Representations},
  2020{\natexlab{a}}.

\bibitem[Wang et~al.(2020{\natexlab{b}})Wang, Liu, Liang, Joshi, and
  Poor]{fednova}
Jianyu Wang, Qinghua Liu, Hao Liang, Gauri Joshi, and H.~Vincent Poor.
\newblock Tackling the objective inconsistency problem in heterogeneous
  federated optimization.
\newblock In H.~Larochelle, M.~Ranzato, R.~Hadsell, M.~F. Balcan, and H.~Lin,
  editors, \emph{Advances in Neural Information Processing Systems}, volume~33,
  pages 7611--7623. Curran Associates, Inc., 2020{\natexlab{b}}.

\bibitem[Wu et~al.(2018)Wu, Xiong, Yu, and Lin]{instance_contrastive_learning}
Zhirong Wu, Yuanjun Xiong, Stella~X. Yu, and Dahua Lin.
\newblock Unsupervised feature learning via non-parametric instance
  discrimination.
\newblock \emph{2018 IEEE/CVF Conference on Computer Vision and Pattern
  Recognition}, pages 3733--3742, 2018.

\bibitem[Yan et~al.(2020)Yan, Misra, Gupta, Ghadiyaram, and
  Mahajan]{clusterfit}
Xueting Yan, Ishan Misra, Abhinav Gupta, Deepti Ghadiyaram, and Dhruv Mahajan.
\newblock Clusterfit: Improving generalization of visual representations.
\newblock In \emph{2020 {IEEE/CVF} Conference on Computer Vision and Pattern
  Recognition, {CVPR} 2020, Seattle, WA, USA, June 13-19, 2020}, pages
  6508--6517. Computer Vision Foundation / {IEEE}, 2020.

\bibitem[Yu et~al.(2006)Yu, Jiang, and Vaidya]{yuji06}
Hwanjo Yu, Xiaoqian Jiang, and Jaideep Vaidya.
\newblock Privacy-preserving svm using nonlinear kernels on horizontally
  partitioned data.
\newblock In \emph{Proceedings of the 2006 ACM Symposium on Applied Computing},
  SAC '06, page 603–610, New York, NY, USA, 2006. Association for Computing
  Machinery.
\newblock ISBN 1595931082.
\newblock \doi{10.1145/1141277.1141415}.

\bibitem[Yu et~al.(2020)Yu, Bagdasaryan, and Shmatikov]{local_adaption}
Tao Yu, Eugene Bagdasaryan, and Vitaly Shmatikov.
\newblock Salvaging federated learning by local adaptation.
\newblock \emph{CoRR}, abs/2002.04758, 2020.

\bibitem[Yurochkin et~al.(2019)Yurochkin, Agarwal, Ghosh, Greenewald, Hoang,
  and Khazaeni]{pfnm}
Mikhail Yurochkin, Mayank Agarwal, Soumya Ghosh, Kristjan Greenewald, Nghia
  Hoang, and Yasaman Khazaeni.
\newblock {B}ayesian nonparametric federated learning of neural networks.
\newblock In Kamalika Chaudhuri and Ruslan Salakhutdinov, editors,
  \emph{Proceedings of the 36th International Conference on Machine Learning},
  volume~97 of \emph{Proceedings of Machine Learning Research}, pages
  7252--7261. PMLR, 2019.

\bibitem[Zhang et~al.(2021{\natexlab{a}})Zhang, Hong, Dhople, Yin, and
  Liu]{fedpd}
Xinwei Zhang, Mingyi Hong, Sairaj Dhople, Wotao Yin, and Yang Liu.
\newblock Fedpd: A federated learning framework with adaptivity to non-iid
  data.
\newblock \emph{IEEE Transactions on Signal Processing}, 69:\penalty0
  6055--6070, 2021{\natexlab{a}}.
\newblock \doi{10.1109/TSP.2021.3115952}.

\bibitem[Zhang et~al.(2021{\natexlab{b}})Zhang, Yang, Yao, Yan, Gonzalez,
  Ramchandran, and Mahoney]{semi-SSFL}
Zhengming Zhang, Yaoqing Yang, Zhewei Yao, Yujun Yan, Joseph~E Gonzalez, Kannan
  Ramchandran, and Michael~W Mahoney.
\newblock Improving semi-supervised federated learning by reducing the gradient
  diversity of models.
\newblock \emph{IEEE International Conference on Big Data (Big Data)},
  2021{\natexlab{b}}.

\bibitem[Zhuang et~al.(2021)Zhuang, Gan, Wen, Zhang, and Yi]{FedU}
Weiming Zhuang, Xin Gan, Yonggang Wen, Shuai Zhang, and Shuai Yi.
\newblock Collaborative unsupervised visual representation learning from
  decentralized data.
\newblock \emph{CoRR}, abs/2108.06492, 2021.

\bibitem[Zhuang et~al.(2022)Zhuang, Wen, and Zhang]{divergenceaware}
Weiming Zhuang, Yonggang Wen, and Shuai Zhang.
\newblock Divergence-aware federated self-supervised learning.
\newblock In \emph{International Conference on Learning Representations}, 2022.

\end{thebibliography}
\bibliographystyle{plainnat}


\newpage
\appendix
\begin{center}
{\bf \Large Supplement for \enquote{Federated Self-supervised Learning for Heterogeneous Clients}}
\end{center}
In this supplementary material, we provide proofs for the key results in the paper in Appendix~\ref{sec:appendix_proof}. 
\section{Proofs}\label{sec:appendix_proof}
We first provide proof of Lemma~\ref{local_epochs} in Appendix~\ref{sec:proof:local_epochs}. Then, the proof of Lemma~\ref{rep_update} is given in Appendix~\ref{sec:proof:rep_update}.
\subsection{Proof of Lemma~\ref{local_epochs}}
\label{sec:proof:local_epochs}
Let $\mathcal{L}_{E}$ denote the loss function after $E$ local epochs. Then, from the Lipschitz smoothness assumption in Assumption~\ref{loss_assum}, we obtain that
\begin{align*}
\mathcal{L}_{1} \leq \mathcal{L}_{0} - \eta  [\nabla \mathcal{L}_{0}^T g_0] + \dfrac{L_1 \eta^2}{2} [||g_0||^2].
\end{align*}
Therefore, we have
\begin{align*}
\E [\mathcal{L}_{1}] & \leq \E[\mathcal{L}_{0}] - \eta \E [\nabla \mathcal{L}_{0}^T g_0] + \dfrac{L_1 \eta^2}{2} \E[||g_0||^2] \\
& = \E[\mathcal{L}_{0}] - \eta ||\nabla \mathcal{L}_{0} ||^2 + \dfrac{L_1 \eta^2}{2} \E[||g_0||^2] \\
& = \E[\mathcal{L}_{0}] - \eta ||\nabla \mathcal{L}_{0} ||^2 + \dfrac{L_1 \eta^2}{2} (Var(g_0) + \E[||g_0||]^2) \\
& \leq \E[\mathcal{L}_{0}] - (\eta - \dfrac{L_1 \eta^2}{2}) ||\nabla \mathcal{L}_{0} ||^2 + \dfrac{L_1 \eta^2}{2} \sigma^2.
\end{align*}
By summing the above bounds over $E$ number of local epochs, we find that
\begin{align*}
\sum_{i=1}^{E} \E[\mathcal{L}_{i}] & \leq \sum_{i=0}^{E-1} \E[\mathcal{L}_{i}] - (\eta - \dfrac{L_1 \eta^2}{2})\sum_{i=0}^{E-1}||\nabla \mathcal{L}_{i} ||^2 + \dfrac{L_1 E \eta^2}{2} \sigma^2 \\
\E[\mathcal{L}_{E}]  & \leq  \mathcal{L}_{0} - (\eta - \dfrac{L_1 \eta^2}{2})\sum_{i=0}^{E-1}||\nabla \mathcal{L}_{i} ||^2 + \dfrac{L_1 E \eta^2}{2} \sigma^2.
\end{align*}
Therefore, if $\eta < \dfrac{2(\sum_{i=0}^{E-1}||\nabla \mathcal{L}_{i} ||^2)}{L_1 (\sum_{i=0}^{E-1} ||\nabla \mathcal{L}_{i} ||^2 + E \sigma^2)}$, we obtain
$\mathcal{L}_{E} \leq  \mathcal{L}_{0}$ and the local loss reduces after $E$ local epochs. As a consequence, we obtain the conclusion of the lemma.

\subsection{Proof of Lemma~\ref{rep_update}}
\label{sec:proof:rep_update}
For any arbitrary client after the global representation update step, if the loss function gets modified to $\loss_{E'}$ from $\loss_{E}$, we have 
\begin{align*}
\loss_{E'} & = \loss_E + \loss_{E'} - \loss_E, \nonumber \\ 
& = \loss_E + \mu \bl \text{d}(.;t) - \text{d}(.;t-1) \br. \nonumber
\end{align*}
As for client $i$, $\text{d}(.;t) = \cka(K_i(t), \Bar{K}(t)) $, we have,
\begin{align*}
\loss_{E'} & = \loss_E + \mu \bl \cka(K_i(t), \Bar{K}(t)) - \cka(K_i(t), \Bar{K}(t-1)) \br.
\end{align*}
Since this holds for all clients, we drop the client index $i$ going forward and obtain that
\begin{align*}
\loss_{E'} & = \loss_E + \mu \bl \text{trace}(K(t) \Bar{K}(t)) - \text{trace}(K(t) \Bar{K}(t-1)) \br \\
& = \loss_E + \mu \bl \text{trace}(K(t) (\Bar{K}(t) - \Bar{K}(t-1))) \br \\
& = \loss_E + \mu \bl \text{trace} (K(t) (\sum_{k=1}^{N} w_k K_k(t) - \sum_{k=1}^{N} w_k K_k(t-1) ) \br \\
& = \loss_E + \mu \bl \sum_{i=1}^{L} \sum_{j=1}^{L} (K(t)_{i,j} (\sum_{k=1}^{N} w_k K_k(t)_{i,j} - \sum_{k=1}^{N} w_k K_k(t-1)_{i,j} ) \br \\
& = \loss_E + \mu \bl \sum_{k=1}^{N} w_k( \sum_{i=1}^{L} \sum_{j=1}^{L} K(t)_{i,j} ( K_k(t)_{i,j} -  K_k(t-1)_{i,j} )) \br \\
& = \loss_E + \mu \bl \sum_{k=1}^{N} w_k( \sum_{i=1}^{L} \sum_{j=1}^{L} K(t)_{i,j} ( \Phi_k(x_i; \W_k^t) . \Phi_k(x_j; \W_k^t) - \Phi_k(x_i; \W_k^{t-1}) . \Phi_k(x_j; \W_k^{t-1}) )) \br. \\
\end{align*}
Given the above equations, we find that
\begin{align*}
\begin{split}
\loss_{E'} = {}& \loss_E + \mu \sum_{k=1}^{N} w_k \sum_{i=1}^{L} \sum_{j=1}^{L} K(t)_{i,j} \bl \Phi_k(x_i; \W_k^{t}) (\Phi_k(x_j; \W_k^{t}) - \Phi_k(x_j; \W_k^{t - 1})) \\
& \hspace{12 em} + \Phi_k(x_j; \W_k^{t - 1})(\Phi_k(x_i; \W_k^{t})  -  \Phi_k(x_i; \W_k^{t-1})). \br 
\end{split}
\end{align*}
Taking norm on both the sides, we get
\begin{align*}
\begin{split}
||\loss_{E'}|| = {}& \biggr|\biggr|\loss_E + \mu \sum_{k=1}^{N} w_k \sum_{i=1}^{L} \sum_{j=1}^{L} K(t)_{i,j} \bl \Phi_k(x_i; \W_k^{t}) (\Phi_k(x_j; \W_k^{t}) - \Phi_k(x_j; \W_k^{t - 1})) \\
& + \Phi_k(x_j; \W_k^{t - 1})(\Phi_k(x_i; \W_k^{t})  -  \Phi_k(x_i; \W_k^{t-1})) \br \biggr|\biggr|
\end{split}\\
\begin{split}
\leq {}& ||\loss_E|| + \mu \biggr|\biggr|\sum_{k=1}^{N} w_k \sum_{i=1}^{L} \sum_{j=1}^{L} K(t)_{i,j} \bl \Phi_k(x_i; \W_k^{t}) (\Phi_k(x_j; \W_k^{t}) - \Phi_k(x_j; \W_k^{t - 1})) \\
& + \Phi_k(x_j; \W_k^{t - 1})(\Phi_k(x_i; \W_k^{t})  -  \Phi_k(x_i; \W_k^{t-1})) \br\biggr|\biggr|
\end{split}\\
\begin{split}
\leq {}& \loss_E + \mu \sum_{k=1}^{N} w_k \sum_{i=1}^{L} \sum_{j=1}^{L} ||K(t)_{i,j}||  \bl ||\Phi_k(x_i; \W_k^{t}) (\Phi_k(x_j; \W_k^{t}) - \Phi_k(x_j; \W_k^{t - 1})) \\
& + \Phi_k(x_j; \W_k^{t - 1})(\Phi_k(x_i; \W_k^{t})  -  \Phi_k(x_i; \W_k^{t-1}))|| \br
\end{split}\\
\begin{split}
\leq {}& \loss_E + \mu \sum_{k=1}^{N} w_k \sum_{i=1}^{L} \sum_{j=1}^{L} ||K(t)_{i,j}|| \bl ||\Phi_k(x_i; \W_k^{t})||. ||(\Phi_k(x_j; \W_k^{t}) - \Phi_k(x_j; \W_k^{t - 1}))|| \\
& + ||\Phi_k(x_j; \W_k^{t - 1})||.||(\Phi_k(x_i; \W_k^{t})  -  \Phi_k(x_i; \W_k^{t-1}))|| \br
\end{split}\\
\leq {}& \loss_E + \mu \sum_{k=1}^{N} w_k \sum_{i=1}^{L} \sum_{j=1}^{L} ||K(t)_{i,j}|| \bl ||\Phi_k(x_i; \W_k^{t})||. ||\text{L}_2 \eta g_{t,k}|| + ||\Phi_k(x_j; \W_k^{t - 1})||.||\text{L}_2 \eta g_{t,k}|| \br. \\
\end{align*}
Taking expectation on both sides of the above bounds, we have
\begin{align*}
\begin{split}
\E[\loss_{E'}] \leq {}& \E[\loss_E] + \mu \sum_{k=1}^{N} w_k \sum_{i=1}^{L} \sum_{j=1}^{L} ||K(t)_{i,j}|| \bl \E[ ||\Phi_k(x_i; \W_k^{t})||. ||\text{L}_2 \eta g_{t,k}||] + \E[||\Phi_k(x_j; \W_k^{t - 1})||.||\text{L}_2 \eta g_{t,k}||] \br
\end{split} \\
\leq {}& \E[\loss_E] + \mu \sum_{k=1}^{N} w_k \sum_{i=1}^{L} \sum_{j=1}^{L} ||K(t)_{i,j}|| \bl R \text{L}_2 \eta P  + R \text{L}_2 \eta P \br \\
= {}& \E[\loss_E] + 2 \mu \eta \text{L}_2 P R \sum_{i=1}^{L} \sum_{j=1}^{L} ||K(t)_{i,j}|| \\
\leq {}& \E[\loss_E] + 2 \mu \eta \text{L}_2 P R \sum_{i=1}^{L} \sum_{j=1}^{L} R^2 \\
\leq {}& \E[\loss_E] + 2 \mu \eta \text{L}_2 P R^3 L^2. \\
\end{align*}
Thus, we have,
\begin{align}
\E[\loss_{E'}] \leq \E[\loss_E] + 2 \mu \eta \text{L}_2 P R^3 L^2, \nonumber
\end{align}
which concludes the proof.

\end{document}